\pdfoutput=1
\documentclass[11pt]{article}

\usepackage[utf8]{inputenc}
\usepackage[T1]{fontenc}
\usepackage[protrusion=true,expansion=false]{microtype}
\usepackage[htt]{hyphenat}
\usepackage{amsmath,amssymb}
\usepackage{booktabs}
\usepackage{multirow}
\usepackage{pgfplots}
\pgfplotsset{compat=1.18}
\usepgfplotslibrary{fillbetween}
\definecolor{figred}{RGB}{227,73,72}
\definecolor{figgraydark}{RGB}{63,72,84}
\definecolor{figgraylight}{RGB}{139,148,161}
\definecolor{mblue}{RGB}{42,120,214}
\definecolor{maqua}{RGB}{27,175,122}
\definecolor{myellow}{RGB}{237,161,0}
\definecolor{morange}{RGB}{235,104,52}
\definecolor{mviolet}{RGB}{74,58,167}
\usepackage{xcolor}
\usepackage{pifont}
\usepackage{listings}
\usepackage[numbers,sort&compress]{natbib}
\usepackage[colorlinks=true,linkcolor=blue!50!black,citecolor=blue!50!black,urlcolor=blue!50!black]{hyperref}
\usepackage[margin=1in]{geometry}

\newcommand{\ztop}{z_{\mathrm{top}}}
\newcommand{\thp}{\theta}
\newcommand{\softmax}{\operatorname{softmax}}
\newcommand{\code}[1]{\texttt{#1}}
\newcommand{\passk}[1]{pass@#1}

\title{\bf Gauge dependence and structured-output corruption\\ in sign-branched repetition penalties\\
\large Measurements across models, inference stacks,\\ and alternative repetition controls}
\author{Peter Hollows}
\date{}

\begin{document}
\maketitle

\begin{abstract}
The multiplicative repetition penalty shipped across the LLM inference ecosystem (HuggingFace,
vLLM, llama.cpp, and a dozen further engines) branches on
the \emph{sign} of each raw logit (divide positives by $\thp$, multiply negatives). But the
softmax is unchanged by adding a constant to every logit, so a model's logit zero-point is
arbitrary (a \emph{gauge} choice), and the sign-branch reads it. Two
measurable consequences follow. (1) The penalty is not well-defined: re-centering a model's logits by a
constant is a provable no-op at $\thp=1$, yet at a routine $\thp=1.3$ it changes $58$--$96\%$ of
greedy tokens, while subtractive and normalized penalties change none; real checkpoints sit at
widely different zero-points, so a fixed
\code{repetition\_penalty} is a different operation on every model. (2) It corrupts
structured output: on $200$ real-world JSON schemas, $\thp=1.3$ drops the rate of valid,
schema-conformant output from $97\%$ to $23\%$. Applying the penalty to normalized
log-probabilities instead of raw logits removes the gauge dependence by construction;
HuggingFace's beam search has applied its processor chain, penalty included, to log-probabilities
since at least v4.0.0, so \code{repetition\_penalty} already names two different operators depending on
decoding strategy. Because equal $\thp$ is not equal strength across the two operators, we also
compare them at \emph{matched} suppression, calibrated per model by search: there the normalized
operator is statistically no worse on any quality metric measured, but on four of seven models
it cannot match the raw operator's suppression at $\thp\ge1.15$, and on six of seven at
$\thp=1.3$, the setting where the corruption was measured. This note gives the mechanism, the measurements (five models up to 7B; two code
models; both effects replicated inside vLLM and llama.cpp through their own samplers), the
per-model calibration map, and the normalized variant.
\end{abstract}

\section{The penalty branches on an unconstrained coordinate}

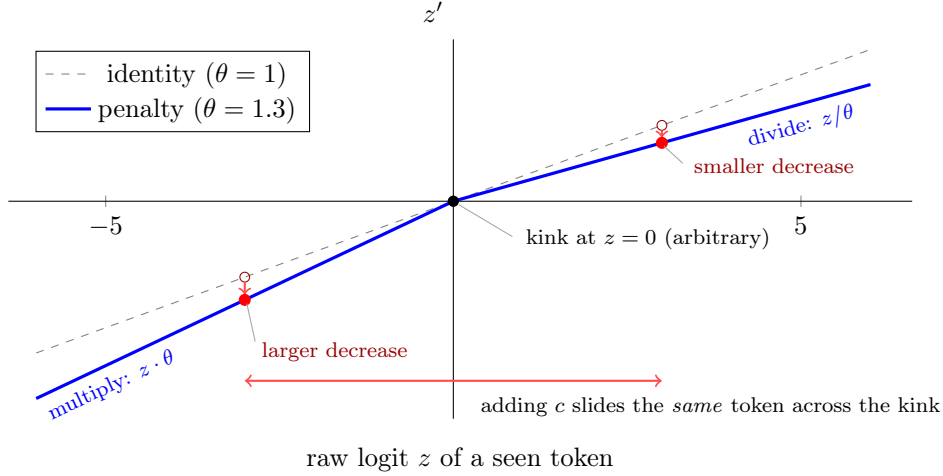
\begin{figure}[t]
\centering
\begin{tikzpicture}
\begin{axis}[
  width=.82\linewidth, height=6.6cm,
  axis lines=middle, axis line style={-},
  xlabel={raw logit $z$ of a seen token}, ylabel={$z'$},
  xmin=-6.4,xmax=6.6, ymin=-8.6,ymax=6.4,
  xtick={-5,0,5}, ytick=\empty,
  x label style={at={(axis description cs:0.5,-0.05)},anchor=north,font=\small},
  y label style={at={(axis description cs:0.47,1.02)},anchor=south,font=\small},
  tick label style={font=\small}, legend style={font=\small,at={(0.03,0.97)},anchor=north west},
  clip=false,
]
\addplot[dashed,gray,domain=-6:6,samples=2]{x}; \addlegendentry{identity ($\thp=1$)}
\addplot[very thick,blue,domain=0:6,samples=2]{x/1.3}
  node[pos=0.82,sloped,below,font=\scriptsize,blue]{divide: $z/\thp$};
\addlegendentry{penalty ($\thp=1.3$)}
\addplot[very thick,blue,domain=-6:0,samples=2]{x*1.3}
  node[pos=0.16,sloped,below,font=\scriptsize,blue]{multiply: $z\cdot\thp$};
\node[circle,draw=red!60!black,inner sep=1.3pt] at (axis cs:3,3) {};
\node[circle,draw=red!60!black,inner sep=1.3pt] at (axis cs:-3,-3) {};
\draw[->,red!70,thick,shorten <=2pt,shorten >=2pt] (axis cs:3,3) -- (axis cs:3,2.308);
\draw[->,red!70,thick,shorten <=2pt,shorten >=2pt] (axis cs:-3,-3) -- (axis cs:-3,-3.9);
\node[circle,fill=black,inner sep=1.5pt] at (axis cs:0,0) {};
\draw[gray!70,thin] (axis cs:0.08,-0.14) -- (axis cs:0.85,-1.4);
\node[font=\scriptsize,anchor=west] at (axis cs:0.9,-1.5) {kink at $z=0$ (arbitrary)};
\node[circle,fill=red,inner sep=1.6pt] at (axis cs:3,2.308) {};
\draw[gray!70,thin] (axis cs:3.03,2.18) -- (axis cs:3.3,1.5);
\node[font=\scriptsize,red!60!black,anchor=west] at (axis cs:3.32,1.4) {smaller decrease};
\node[circle,fill=red,inner sep=1.6pt] at (axis cs:-3,-3.9) {};
\draw[gray!70,thin] (axis cs:-2.97,-4.05) -- (axis cs:-2.85,-5.55);
\node[font=\scriptsize,red!60!black,anchor=west] at (axis cs:-2.9,-5.95) {larger decrease};
\draw[<->,red!70,thick] (axis cs:-3,-7.1) -- (axis cs:3,-7.1);
\node[font=\scriptsize,anchor=north west] at (axis cs:0.25,-7.35)
  {adding $c$ slides the \emph{same} token across the kink};
\end{axis}
\end{tikzpicture}
\caption{The penalty maps a seen token's logit $z$ to $z'$: positive logits are divided by $\thp$
(shallow slope), negative logits multiplied (steep slope), meeting at a kink at $z=0$. Open marks
show each token's unpenalized value on the identity line; the arrows show the penalty's decrease,
smaller on the divide branch and larger on the multiply branch. Because the softmax does
not change when a constant is added to every logit, the model can slide every seen token left or
right across this kink without changing any output probability, so the \emph{same} token can be divided under one gauge and
multiplied under another and receive a different penalty.}
\label{fig:kink}
\end{figure}

A repetition penalty is supposed to be a clean scalar knob: set \code{repetition\_penalty}
$\approx 1.1$--$1.3$, get less repetition, and expect it to mean roughly the same thing on any
model. The multiplicative form used almost everywhere (HuggingFace's
\code{RepetitionPenaltyLogitsProcessor}, vLLM's \code{repetition\_penalty}, llama.cpp's
\code{repeat\_penalty}, all inherited from CTRL \citep{keskar2019ctrl}) does not have that
property; Section~\ref{sec:stacks} surveys how widely the operator ships.\footnote{Code, frozen
pre-registrations, and raw outputs for every measurement in this note are at
\href{https://github.com/captainpete/repetition-penalty-gauge}{\urlstyle{same}\nolinkurl{github.com/captainpete/repetition-penalty-gauge}}
--- ``the repository'' throughout.} It
transforms an already-seen token's logit $z_i$ by
\begin{equation}\label{eq:op}
z_i' = \begin{cases} z_i/\thp & z_i \ge 0 \quad(\text{divide}),\\ z_i\cdot\thp & z_i < 0
\quad(\text{multiply}),\end{cases}
\end{equation}
i.e.\ it branches on the \emph{sign} of the raw logit. The verbatim implementation is three lines:
\begin{lstlisting}[language=Python]
score = torch.gather(scores, 1, input_ids)
score = torch.where(score < 0, score * penalty, score / penalty)  # sign-branch
scores_processed = scores.scatter(1, input_ids, score)
\end{lstlisting}
The sign-branch is itself a fix: the naive version divided \emph{every} seen logit by $\thp$, but
dividing a negative logit by $\thp>1$ moves it toward zero and \emph{raises} the token's
probability, which is backwards; this was reported to HuggingFace in 2019 \citep{hf2302}, and the
sign-branch is the accepted remedy. The problem is one level up. The softmax obeys
$\softmax(z + c) = \softmax(z)$ for any scalar $c \in \mathbb{R}$, where $z + c$ adds $c$ to
every coordinate of the logit vector. A uniform shift therefore leaves the next-token
distribution unchanged, so the greedy token is unchanged exactly, and sampled output is
unchanged token-for-token under a fixed random seed. We call a choice of the constant $c$ a
\emph{gauge}: the model's behavior fixes the distribution but not the gauge. So the location of the sign-branch's kink,
$z_i=0$, is a coordinate not constrained by training: the loss depends on the logits only through
the softmax. The accepted fix for the naive penalty branches on this unconstrained coordinate
(Figure~\ref{fig:kink}). Sections~2 and~3 measure the two consequences, labeled A1 and A2
after their pre-registration identifiers.

\section{Consequence 1: the penalty is not well-defined (A1)}

Where a model's logits sit relative to zero differs widely across real checkpoints:
measured at decode time on five checkpoints (Table~\ref{tab:zeropoint}), the fraction of
seen-token logits that are positive, and so \emph{divided} rather than multiplied, ranges from
$0.17$ on gpt2 to $0.95$ on Qwen2.5-Coder-7B. The same \code{repetition\_penalty=1.3} therefore
takes the multiply branch for most seen tokens on the first model and the divide branch for
nearly all on the second. A fixed \code{repetition\_penalty} is a different operation on every
model.

Within one model this becomes a controlled experiment. Adding a constant $c$ to every logit has no
effect on the model's output distribution (Section~1), so we decode the \emph{same} model twice at
the same $\thp$, once with $c=+5$ and once with $c=-5$ added to the logits (equivalently,
$\code{lm\_head.bias}\mathrel{+}=c$), and count the fraction of greedy positions where the two
runs differ: the flip rate. For a well-defined intervention the flip rate is zero, and for the
other standard repetition controls it is: with a subtractive presence penalty ($z_i - \alpha$ on
seen tokens, $\alpha=1$) the two runs are token-for-token identical on every model, and the same
holds for the multiplicative penalty applied to normalized log-probabilities (Section~\ref{sec:fix}).
Prompts follow the standard open-ended-generation protocol: $200$ 32-token prefixes sampled from
the WikiText-103 test set \citep{merity2017pointer,su2022contrastive}, $200$ greedy tokens each,
i.e.\ $40{,}000$ compared positions per model. At $\thp=1$ the CTRL runs are also identical (a
provable no-op). At $\thp=1.3$ the CTRL operator flips $58$--$96\%$ of greedy tokens
(Table~\ref{tab:a1}); this holds at 7B and on the base/RLHF pair tested. The probe needs no synthetic
constant either: re-centering each model by its own median logit (a gauge it could equally have
shipped) flips $71$--$97\%$ (Table~\ref{tab:zeropoint}, last column). The own-median offsets are
larger than the $\pm5$ probe (for gpt2, $161$ logit units), which is why gpt2's $0.71$ there
exceeds its $0.58$ in Table~\ref{tab:a1}.

\begin{table}[t]
\centering
\caption{Where five real checkpoints sit relative to the penalty's sign boundary, at decode time.
This set spans the divide-fraction range and includes the two code models used in Consequence 2;
the controlled flip experiment (Table~\ref{tab:a1}) uses the five base/RLHF models. The
divide-fraction spans $0.17$--$0.95$, and re-centering each model by its own median logit (a
behaviorally invisible gauge) flips most of its greedy tokens at $\thp=1.3$. gpt2's per-position
logits are bimodal (deciles of the per-position top-1 logit span roughly $-230$ to $+150$), so a large negative median
coexists with a positive divide-fraction.}
\label{tab:zeropoint}
\setlength{\tabcolsep}{4pt}
\begin{tabular}{lccc}
\toprule
Model & frac.\ seen logit $>0$ (divided) & median top-1 logit & flip @ own-median gauge \\
\midrule
gpt2              & $0.17$ & $-161.4$ & $0.71$ \\
gpt2-large        & $0.52$ & $13.3$  & $0.96$ \\
starcoder2-7b     & $0.78$ & $18.8$  & $0.96$ \\
pythia-2.8b       & $0.88$ & $19.7$  & $0.96$ \\
Qwen2.5-Coder-7B  & $0.95$ & $24.8$  & $0.97$ \\
\bottomrule
\end{tabular}
\end{table}

\begin{table}[t]
\centering
\caption{Gauge flip rate: fraction of greedy tokens that differ between $c=+5$ and $c=-5$, two
configurations with identical output distributions. A well-defined intervention flips nothing;
the subtractive presence penalty and the normalized multiplicative penalty flip none of the
$40{,}000$ positions on any model, while the CTRL sign-branch flips most tokens. The
$\thp{=}1$ column is the harness control: any implementation error would flip tokens there.}
\label{tab:a1}
\begin{tabular}{lcccc}
\toprule
Model & CTRL, $\thp{=}1$ (gate) & CTRL, $\thp{=}1.3$ & subtractive, $\alpha{=}1$ & normalized \\
\midrule
gpt2-large           & $0.000$ & $0.96$ & $0.000$ & $0.000$ \\
pythia-2.8b          & $0.000$ & $0.94$ & $0.000$ & $0.000$ \\
gpt2                 & $0.000$ & $0.58$ & $0.000$ & $0.000$ \\
Qwen2.5-7B (base)    & $0.000$ & $0.92$ & $0.000$ & $0.000$ \\
Qwen2.5-7B-Instruct  & $0.000$ & $0.92$ & $0.000$ & $0.000$ \\
\bottomrule
\end{tabular}
\end{table}

\begin{figure}[t]
\centering
\begin{tikzpicture}
\begin{axis}[
  name=f2a, width=.425\linewidth, height=6.4cm,
  xlabel={position in continuation},
  ylabel={likelihood cost vs unpenalized (nats)},
  xmin=0, xmax=205, ymin=-8, clip=false, ytick={0,50,100,150,200},
  title style={font=\small}, title={(a) cost under two gauges (gpt2-large)},
  label style={font=\small}, tick label style={font=\small},
  legend style={font=\scriptsize, at={(0.03,0.97)}, anchor=north west, draw=none, fill=none},
  legend cell align=left, axis lines=left, grid=major, grid style={gray!15},
]
\addplot[figred, very thick] table[x=pos,y=shipped]{figdata/fig2a.dat};
\addlegendentry{CTRL, as shipped ($c{=}0$)}
\addplot[figred, very thick, dashed] table[x=pos,y=median]{figdata/fig2a.dat};
\addlegendentry{CTRL, own median ($c{=}{-}13.3$)}
\addplot[figgraydark, thick] table[x=pos,y=sub]{figdata/fig2a.dat};
\addlegendentry{subtractive $\alpha{=}1$}
\addplot[figgraylight, thick, densely dotted] table[x=pos,y=norm]{figdata/fig2a.dat};
\addlegendentry{normalized $\thp{=}1.3$}
\draw[<->, figred, thick] (axis cs:201,232) -- (axis cs:201,60);
\node[font=\scriptsize, align=right, text=figred, anchor=east] at (axis cs:195,112)
  {same knob,\\$4.1\times$ the cost};
\end{axis}
\begin{axis}[
  at={(f2a.south east)}, xshift=1.35cm, anchor=south west,
  width=.425\linewidth, height=6.4cm,
  xlabel={position in continuation},
  ylabel={$|\Delta$ cum.\ log-prob$|$ (nats)},
  xmin=0, xmax=205, ymin=-8, ymax=290, clip=false, ytick={0,50,100,150,200,250},
  title style={font=\small}, title={(b) gauge gap of the effect, $c=\pm5$},
  label style={font=\small}, tick label style={font=\small},
  legend style={font=\scriptsize, at={(0.03,0.97)}, anchor=north west, draw=none, fill=white,
    fill opacity=0.8, text opacity=1},
  legend cell align=left, axis lines=left, grid=major, grid style={gray!15},
]
\addplot[name path=t1, draw=none, forget plot] table[x=pos,y=q3]{figdata/fig2b_gpt2.dat};
\addplot[name path=b1, draw=none, forget plot] table[x=pos,y=q1]{figdata/fig2b_gpt2.dat};
\addplot[mblue, opacity=0.12, forget plot] fill between[of=t1 and b1];
\addplot[mblue, thick, forget plot] table[x=pos,y=med]{figdata/fig2b_gpt2.dat};
\addplot[name path=t2, draw=none, forget plot] table[x=pos,y=q3]{figdata/fig2b_gpt2large.dat};
\addplot[name path=b2, draw=none, forget plot] table[x=pos,y=q1]{figdata/fig2b_gpt2large.dat};
\addplot[maqua, opacity=0.12, forget plot] fill between[of=t2 and b2];
\addplot[maqua, thick, forget plot] table[x=pos,y=med]{figdata/fig2b_gpt2large.dat};
\addplot[name path=t3, draw=none, forget plot] table[x=pos,y=q3]{figdata/fig2b_pythia.dat};
\addplot[name path=b3, draw=none, forget plot] table[x=pos,y=q1]{figdata/fig2b_pythia.dat};
\addplot[myellow, opacity=0.12, forget plot] fill between[of=t3 and b3];
\addplot[myellow, thick, forget plot] table[x=pos,y=med]{figdata/fig2b_pythia.dat};
\addplot[name path=t4, draw=none, forget plot] table[x=pos,y=q3]{figdata/fig2b_qwen.dat};
\addplot[name path=b4, draw=none, forget plot] table[x=pos,y=q1]{figdata/fig2b_qwen.dat};
\addplot[morange, opacity=0.12, forget plot] fill between[of=t4 and b4];
\addplot[morange, thick, forget plot] table[x=pos,y=med]{figdata/fig2b_qwen.dat};
\addplot[name path=t5, draw=none, forget plot] table[x=pos,y=q3]{figdata/fig2b_qwenI.dat};
\addplot[name path=b5, draw=none, forget plot] table[x=pos,y=q1]{figdata/fig2b_qwenI.dat};
\addplot[mviolet, opacity=0.12, forget plot] fill between[of=t5 and b5];
\addplot[mviolet, thick, forget plot] table[x=pos,y=med]{figdata/fig2b_qwenI.dat};
\addlegendimage{mblue, line width=1.4pt}\addlegendentry{gpt2}
\addlegendimage{maqua, line width=1.4pt}\addlegendentry{gpt2-large}
\addlegendimage{myellow, line width=1.4pt}\addlegendentry{pythia-2.8b}
\addlegendimage{morange, line width=1.4pt}\addlegendentry{Qwen2.5-7B}
\addlegendimage{mviolet, line width=1.4pt}\addlegendentry{Qwen2.5-7B-Instruct}
\addlegendimage{black!75, line width=1.8pt}\addlegendentry{controls}
\draw[black!75, very thick] (axis cs:0,0) -- (axis cs:200,0);
\end{axis}
\end{tikzpicture}
\caption{The cost of each intervention in model log-likelihood: traces scored
token-by-token under the unpenalized distribution, mean over the $200$ WikiText-103 prefixes.
(a)~gpt2-large under two equally legitimate gauges of the same checkpoint (as shipped, and
re-centered by its own median logit): the same $\thp=1.3$ costs $4.1\times$ more in one gauge
than the other, and the two prices straddle the subtractive penalty's. The controls each have a
single, gauge-independent cost curve. (b)~Cumulative log-probability gap between the $c=\pm5$
gauge runs (median and IQR over prefixes): it grows without saturating on every model, while the
two control operators (the subtractive presence penalty, $\alpha=1$, and the normalized penalty)
sit at exactly zero at every position.}
\label{fig:cost}
\end{figure}
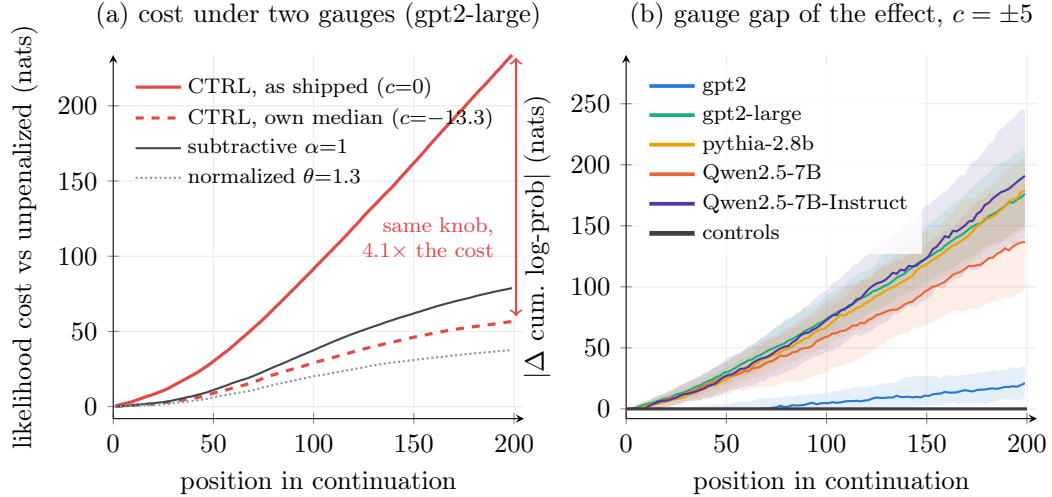

\paragraph{The likelihood cost of the intervention.} Scoring every trace
token-by-token under the model's unpenalized distribution prices each intervention in
log-likelihood (Figure~\ref{fig:cost}). For
the controls that price is well-defined: one deterministic cost curve each. For
the CTRL operator it is not: on gpt2-large at $\thp=1.3$, the model as shipped pays $235$ nats
over $200$ tokens, and the same model re-centered by its own median logit pays $57$; the
subtractive penalty's $79$ lies between the two. Across models, the cumulative
log-likelihood gap between the $c=\pm5$ gauge runs grows roughly linearly, reaching $140$--$190$
nats by position $200$ on four of five models (Figure~\ref{fig:cost}b); for the controls it is
zero at every position. The settings are routine values, not calibrated to each other:
$\thp=1.3$ is the setting tested throughout, and $\alpha=1$ is mid-range of the hosted-API
presence-penalty interval. Cost levels across \emph{different} operators therefore are not
comparable; the comparisons that do not depend on these choices are between the two CTRL gauge
runs, and of each control against zero.

\paragraph{Reading the flip rate.} A flip is not by itself a quality claim: a penalty is supposed
to change tokens, and open-ended prose has no single correct continuation. The controls fix the
number's meaning. Up to
the first divergence the two runs share an identical context, so the first differing token cannot
be inherited from earlier differences; it can only come from the gauge changing that one greedy
decision. The control operators run the same 200-token greedy process without one such flip in
$40{,}000$ positions. Under the CTRL operator, divergence reaches most prefixes (200/200 on four models, 180/200 on gpt2), with median first
divergence at position 4--10 (62 on gpt2); once diverged, two continuations stay
$\sim$$99\%$ different, as any two different texts do, so the headline rate reflects early and
pervasive seeding rather than a per-position disagreement rate. The gauge also sets how strongly
the penalty acts: the size of the push depends on where a seen logit sits relative to the kink
(Figure~\ref{fig:kink}), so the same $\thp$ can over- or under-suppress depending on the
zero-point. \code{repetition\_penalty=1.3} does not name one
behavior; the default penalty is an inference-time intervention whose effect depends in part on
the logit zero-point, a coordinate training sets only incidentally. Where correctness is
defined, the flips stop being neutral: that is Consequence 2.

\section{Consequence 2: it corrupts structured output (A2)}

Structured output has repetition the grammar \emph{requires}: JSON must reclose \code{\}}, reseparate
with \code{,}, requote with \code{"}; code must re-indent. Those delimiters have appeared before, so
the penalty pushes them down, and it cannot tell ``the grammar needs this'' from ``the model is
looping.'' At a delimiter where the correct token is the confident top choice (logit $\ztop>0$) with
the runner-up a gap $g$ below (and itself unpenalized; positions where both are penalized obey a
two-sided variant), the penalty ($\ztop\mapsto\ztop/\thp$) flips the greedy choice to the
runner-up exactly when
\begin{equation}\label{eq:thr}
\ztop/\thp < \ztop - g \iff g < \ztop\,(1-1/\thp).
\end{equation}
On StarCoder2-7B, decoding the $164$ HumanEval prompts \citep{chen2021humaneval}, this closed
form predicts the observed penalty-induced flips with balanced
accuracy $0.999$ over the $48{,}919$ positions where its assumptions hold (top token penalized on
the divide branch, runner-up unpenalized; pooled over $\thp\in\{1.1,\dots,1.5\}$), and all
$27{,}720$ flips there land on the pre-penalty
runner-up (this replicates on
Qwen2.5-Coder-7B). End to end: generating complete JSON objects against $200$ real-world schemas
sampled from JSONSchemaBench \citep{geng2025jsonschemabench}, a routine
$\thp=1.3$ takes the rate of parseable, schema-valid output from $97\%$ to $23\%$
(Table~\ref{tab:fix}); at $\thp=1.1$ (Ollama's default, Section~\ref{sec:stacks}) the rate is
$95.5\%$ ($191/200$; $97.5\%$ under the normalized operator at the same $\thp$), so the collapse
lies between $1.1$ and $1.3$. The surviving valid outputs at $\thp=1.3$ are dominated by
near-empty schemas with no required properties, i.e.\ instances with few grammar-obligatory
delimiters to corrupt.\footnote{We also pre-registered, and our data refuted, two
sharper claims: that corruption hits the \emph{most confident} tokens first (it is gap-driven, so it
hits \emph{less} confident tokens) and that it is strongly code-specific (the code-vs-prose ratio is
real but model-dependent, $2.3\times$ on StarCoder2 vs $5.6\times$ on Qwen). Details and the full
audit are in the repository.}

\section{The operator ships across the inference ecosystem}\label{sec:stacks}

Both effects replicate inside vLLM and llama.cpp. The main experiments use the HuggingFace
implementation; we re-ran A1 and A2 inside both other stacks, on the same benchmark inputs,
through each stack's own sampler and its own
knob: vLLM 0.8.5.post1 (pinned by the test host's driver; the operator in \code{vllm/main} at
\code{e196268} is mathematically identical, so the pin does not narrow the finding) and llama.cpp
at master commit \code{4fc4ec55}. The A1
gauge probe passes its validity gate in both stacks (exactly $0$ of $40{,}000$ greedy tokens flip at
$\thp=1$) and reproduces the divergence at $\thp=1.3$ on gpt2-large: flip rate $0.964$ in both,
matching HuggingFace, and the vLLM run is token-identical to the HuggingFace run at every compared
position. The A2 JSON task (Qwen2.5-Coder-7B, same
schemas and validator) falls
from $0.97$ valid at $\thp=1$ to $0.24$ (vLLM) and $0.29$ (llama.cpp) at $\thp=1.3$ (HF: $0.23$).
llama.cpp's default 64-token penalty window makes this worse, not better ($0.12$ at $\thp=1.3$):
on multi-hundred-token structured outputs the sliding window concentrates the penalty on the most
recent, grammar-obligatory tokens.\footnote{Measuring inside llama.cpp also
surfaced an unrelated defect: \code{--repeat-last-n -1} is documented as ``context size'' but the
sampler clamps the value to $0$, silently disabling the penalty
(\code{src/llama-sampler.cpp}, \code{std::max(penalty\_last\_n, 0)}); reported as
\href{https://github.com/ggml-org/llama.cpp/issues/25388}{ggml-org/llama.cpp\#25388}.} Scripts and raw outputs are
in the repository (\code{code/smoke\_*}, \code{results/smoke\_*\_bench}).

The operator is not confined to those three. A source survey finds it in twelve
further engines (TGI, SGLang, TensorRT-LLM, ExLlamaV2, mlx-lm, LMDeploy, aphrodite-engine,
KoboldCpp, mistral.rs, candle, text-generation-webui, and Ollama), in every case a sign-branch on
raw logits with no normalization; Ollama enables the penalty by default
(\code{repeat\_penalty} $= 1.1$). These engines share the operator's form and with it both
effects; we have not
run the probes inside them (Table~\ref{tab:stacks}). A git-history trace in the repository shows
the sign-branch was reached independently at least twice, in HuggingFace's 2019 fix for the naive
divide and in an unrelated 2023 re-derivation in llama.cpp from the CTRL paper, while the surveyed
engines each copied a prior implementation rather than deriving it anew; neither derivation records
the gauge dependence. The closest prior observation we found is a 2023 llama.cpp issue noting the
kink's conflict with softmax shift-invariance, closed without action \citep{llamacpp2970}.
Because each engine keeps its own copy rather than importing a shared one, no
single upstream patch reaches them all.

\begin{table}[t]
\centering
\caption{Penalty form by stack. The multiplicative CTRL form (sign-branch on raw logits) is subject
to both effects; the subtractive presence/frequency form subtracts a constant, is
shift-invariant, and is immune. The three multiplicative stacks are measured directly (HuggingFace
in the main experiments; vLLM and llama.cpp in this section); the surveyed row covers the twelve
engines named in the text, classified by operator form, not probed directly.}
\label{tab:stacks}
\setlength{\tabcolsep}{3.5pt}\small
\begin{tabular}{llccc}
\toprule
Stack / knob & Penalty form & Normalizes first? & A1? & A2? \\
\midrule
HF \code{repetition\_penalty}        & multiplicative (CTRL) & no (default) & yes & yes \\
llama.cpp \code{repeat\_penalty}     & multiplicative (CTRL) & no & yes & yes \\
vLLM \code{repetition\_penalty}      & multiplicative (CTRL) & no & yes & yes \\
twelve surveyed engines              & multiplicative (CTRL) & no & by form & by form \\
Hosted-API \code{presence\_penalty}  & subtractive (constant) & n/a & no & no \\
\code{frequency\_penalty} (API/vLLM) & subtractive (count-scaled) & n/a & no & no \\
\bottomrule
\end{tabular}
\end{table}

\section{Normalizing before the penalty}\label{sec:fix}

Both problems have one root, branching on the sign of an un-normalized logit. Shift-invariant
repetition controls already exist (the subtractive family of Table~\ref{tab:stacks}); this
section measures what changes if the multiplicative operator itself reads a shift-invariant
coordinate, i.e.\ the penalty applied to $\ell = \log\softmax(z)$ instead of to
$z$. Two things happen. (a)~$\log\softmax$ is
itself shift-invariant, so the arbitrary gauge $c$ is gone and A1 disappears by construction.
(b)~Every $\ell_i \le 0$, so the sign-branch stops branching: every seen token takes the multiply
path, $\ell_i \mapsto \thp\,\ell_i$. Since $\thp\log p_i = \log p_i^{\thp}$, this is
$p_i \mapsto p_i^{\thp}$ in probability space: the normalized penalty tempers
each seen token's probability downward by a common power. It is a monotone, well-defined suppression that
reads a coordinate the model's distribution determines (where $\sum_i e^{\ell_i}=1$) rather than
one the training objective leaves free. HuggingFace's beam search has
applied its entire logits-processor chain, the repetition penalty included, to $\log\softmax$
output since at least \code{transformers} v4.0.0 (2020), while greedy and sampled decoding apply
the raw form.\footnote{\code{src/transformers/generation/utils.py} (\code{generation\_utils.py} in
v4.0.0): in \code{\_beam\_search}, \code{log\_softmax} precedes the processor call; the sampling
path passes raw logits. The ordering is present from v4.0.0's \code{beam\_search} (and already in
v3.5.1's \code{\_generate\_beam\_search}) through \code{main} at \code{a5f41e09}.} The same
\code{repetition\_penalty} thus already names both operators within one library, selected by
decoding strategy.

At equal $\thp$, normalizing first removes both effects (Table~\ref{tab:fix}): the A1 gauge
flip rate falls to $0$ at every $\thp$ on every model, delimiter corruption drops $\sim$$80\times$, and
JSON validity at $\thp=1.3$ returns from $23\%$ to $97\%$, its $\thp=1$ level. It still breaks
degenerate loops (repetition rate falls monotonically in $\thp$). But equal $\thp$ is not equal
strength: the normalized operator acts on log-probabilities ($O(1)$) rather than raw logits
($O(10)$), so at the same $\thp$ it also suppresses repetition less, and an equal-$\thp$ reading
conflates the change of operator with a change of suppression. The remainder of this section
holds suppression fixed.

\paragraph{Matching suppression requires a per-model $\thp'$.} For each raw setting $\thp$ (the
\emph{anchor}), we search by bisection (cap $\thp'\le10$) for the normalized $\thp'$ that
reproduces the raw operator's greedy repetition rate (the fraction of generated tokens already
seen) on the $16$ fixed prefixes of the pre-registered runs, $256$ tokens each. Seven
models: the four of Table~\ref{tab:calib}, which continue to the matched-suppression comparison
below, plus three checkpoints in the repository. Ollama's default $\thp=1.1$ is matchable on six
of the seven, at $\thp'$ from
$1.74$ (gpt2-large, in the extended set) to $6.63$. So $\thp'$ must be calibrated per model; no
conversion constant exists.

The map also ends: the normalized
curve plateaus, leaving raw suppression at $\thp\ge1.15$ unreachable on four of the seven models
($\thp=1.3$ on all but gpt2-large; on gpt2, everything above $1.02$). Matching is on repetition
rate, the primary of three frozen metrics (the others: longest run and distinct-2); under the
others the map shifts, so a tabulated
$\thp'$ is a verified match, not a unique one. The ceiling has a closed form: a greedy flip of
a seen top token requires $p_{\mathrm{top}}^{\thp'} < p_{\mathrm{run}}$, i.e.\
$\thp' > \log p_{\mathrm{run}} / \log p_{\mathrm{top}}$, which diverges as
$p_{\mathrm{top}}\to1$; on gpt2's loops the bound sits in the thousands. Why the ceiling
binds on some models and not others is open: we pre-registered loop confidence as the
cross-model predictor, and the test refuted it (verdict in the repository).

\begin{table}[t]
\centering
\caption{Normalize-before-penalize at \emph{equal} $\thp$. Applying the penalty to
$\log\softmax(z)$ removes the A1 gauge flips, reduces delimiter corruption $\sim$$80\times$, and
returns JSON validity to its $\thp=1$ level. The delimiter-flip rate counts flips landing on
structural tokens as a fraction of all generated code tokens (mean over generations, $\thp$
pooled over $1.1$--$1.5$); it is a different population from the formula-exact positions of
Section~3. Equal-$\thp$ readings conflate operator and suppression: at the same $\thp$ the
normalized operator also suppresses less.}
\label{tab:fix}
\begin{tabular}{lcc}
\toprule
Metric & raw logits (default) & normalized \\
\midrule
Gauge flip rate @$\thp{=}1.3$ (gpt2-large / pythia / gpt2) & $0.96$ / $0.94$ / $0.58$ & $0.00$ / $0.00$ / $0.00$ \\
Delimiter-flip rate (code, StarCoder2-7B)              & $0.139$ & $0.0017$ \\
JSON schema-valid rate @$\thp{=}1.1$ (Qwen2.5-Coder-7B) & $0.955$ & $0.975$ \\
JSON schema-valid rate @$\thp{=}1.3$ (Qwen2.5-Coder-7B) & $0.23$ & $0.97$ \\
\bottomrule
\end{tabular}
\end{table}

\begin{table}[t]
\centering
\caption{Calibration map: the normalized $\thp'$ that matches the raw operator's greedy
repetition rate at each anchor $\thp$, by bisection with ceiling $\thp'\le10$
(\ding{55}~$=$ unreachable: the normalized operator's repetition-rate floor exceeds the raw
target). Shown: the four models of the matched-suppression comparison; three further
checkpoints extend the map in the repository.}
\label{tab:calib}
\begin{tabular}{lcccc}
\toprule
Anchor $\thp$ (raw) & gpt2 & pythia-2.8b & Qwen2.5-7B & Qwen2.5-Coder-7B \\
\midrule
$1.02$ & $2.86$ & $1.21$ & $1.28$ & $1.28$ \\
$1.05$ & \ding{55} & $1.70$ & $1.84$ & $3.25$ \\
$1.08$ & \ding{55} & $2.20$ & $3.25$ & $4.38$ \\
$1.10$ & \ding{55} & $2.51$ & $5.50$ & $6.63$ \\
$1.15$ & \ding{55} & $3.81$ & \ding{55} & \ding{55} \\
$1.20$ & \ding{55} & $6.63$ & \ding{55} & \ding{55} \\
$1.30$ & \ding{55} & \ding{55} & \ding{55} & \ding{55} \\
\bottomrule
\end{tabular}
\end{table}

\paragraph{At matched suppression.} With suppression held equal at the calibrated pairs
(verified per pair; pythia-2.8b at the deployed anchor runs $0.510$ raw vs $0.509$ normalized),
the comparison is unconfounded (Table~\ref{tab:matched}). Under $c=\pm5$ the raw operator
flips $64$--$93\%$ of greedy tokens across the ten calibrated $(\thp,\thp')$ pairs (one
matchable anchor on gpt2, three each on the other three models; Table~\ref{tab:matched} shows
the deployed ones). The normalized operator, shift-invariant, flips
none (checked numerically on the $16$ calibration prefixes). gpt2 flips $0.71$ already at its
$\thp=1.02$ pair, matching its own-median rate at $\thp=1.3$ (Table~\ref{tab:zeropoint}): its
logits are large enough that even a $2\%$ change flips early tokens, and early flips decide the
rest (Section~2). On
canonical HumanEval (Qwen2.5-Coder-7B; $\thp=1$ baseline $0.59$ under both operators, consistent
with the published figure), quality follows suppression, not operator. Re-measuring suppression
on the generated code itself (completions shorter than $8$ tokens are excluded from the rate:
$10$ of $164$ at $\thp=1$, $23$--$24$ at the deployed pair), the two gentler pairs stay matched (gaps $\le0.009$) and tie on
\passk{1}. At the deployed pair the match fails on code: $\thp'=6.63$ delivers three-quarters of
raw's suppression there (repetition rate $0.524$ vs $0.492$, from a $0.618$ unpenalized
baseline), and \passk{1} splits accordingly:
raw $0.470$, normalized $0.530$, a $6.1$-point gap ($95\%$ CI $[{+}1.2, {+}11.6]$). The gap
conflates operator with suppression: the prose-calibrated pair suppresses code less, and less
suppression does less damage. At equal suppression, measured quality is a tie everywhere. Open-ended quality
(distinct-$n$, judge-scored; $16$ prompts; per-pair numbers in the repository) shows no
significant difference at any pair; with $16$ prompts this is a low-powered tie rather than
demonstrated equivalence. JSON-schema validity (six schemas,
$48$ instances) is $1.00$ for both operators at every matched pair: the raw operator's collapse
occurs at $\thp\approx1.3$, which the normalized operator cannot reach. It avoids A2's collapse
through its ceiling rather than by rescuing that regime, consistent with the equal-$\thp$
reading of Table~\ref{tab:fix}.

\begin{table}[t]
\centering
\caption{Raw vs.\ normalized at matched suppression (the calibrated pairs of
Table~\ref{tab:calib}; gpt2 at its one matchable anchor $1.02$, others at the deployed $1.10$).
Gauge flip rate as in Table~\ref{tab:a1}, on the $16$ calibration prefixes; HumanEval \passk{1}
with the canonical harness; --- $=$ not measured. At matched suppression the normalized operator
is statistically no worse on any measured metric; the deployed-anchor \passk{1} gap favors it
but conflates operator with suppression level (see text).}
\label{tab:matched}
\setlength{\tabcolsep}{4.5pt}
\begin{tabular}{lcccc}
\toprule
Model (pair $\thp\!\to\!\thp'$) & flip, raw & flip, norm. & \passk{1}, raw & \passk{1}, norm. \\
\midrule
gpt2 ($1.02\!\to\!2.86$)              & $0.71$ & $0.00$ & --- & --- \\
pythia-2.8b ($1.10\!\to\!2.51$)       & $0.93$ & $0.00$ & --- & --- \\
Qwen2.5-7B ($1.10\!\to\!5.50$)        & $0.87$ & $0.00$ & --- & --- \\
Qwen2.5-Coder-7B ($1.10\!\to\!6.63$)  & $0.91$ & $0.00$ & $0.470$ & $0.530$ \\
\bottomrule
\end{tabular}
\end{table}

HuggingFace also ships this normalization as a standalone processor
(\code{LogitNormalization}); in the sampling path it is off by default and, when enabled, runs
after the penalty; the difference is only the ordering. The same reorder applies to every stack
carrying the multiplicative form (Section~\ref{sec:stacks}), with $\thp'$ re-calibrated per
model (Table~\ref{tab:calib}). Within its ceiling, the
normalized operator is the gauge-correct form of the $\thp$-style knob; for stronger
suppression, the subtractive family is the shift-invariant option that reaches confident tokens
(a constant subtraction moves every logit equally). Standardizing on the subtractive family, or
on a redesigned penalty such as the LZ penalty \citep{ginart2025lz}, avoids the multiplicative
operator entirely.

\section{Scope}

We study the multiplicative CTRL operator of Eq.~\eqref{eq:op}; subtractive penalties are immune and
stacks that already normalize before penalizing are exempt. A1 is measured on five models up to 7B
(base and RLHF) over $200$ WikiText-103 test prefixes, with subtractive and normalized penalties
as measured controls; A2 on two 7B code models over the $164$
HumanEval prompts and $200$ JSONSchemaBench schemas; both replicate inside vLLM and llama.cpp
through each
stack's own sampler, on the same inputs. The matched-suppression comparison of
Section~\ref{sec:fix} verifies each pair's match before comparing (on code, re-measured on the
generations themselves) and scores code quality with the canonical HumanEval harness (its
$\thp=1$ baseline is consistent with the model's published \passk{1} under both operators). The original pre-registered
runs used $16$ fixed prompts ($18$ for A2)
and $6$ hand-written schemas; their frozen decision rules, verdicts, and raw outputs remain in the
repository, and their verdicts are consistent with the benchmark numbers reported here. All
experiments are inference-only, greedy, on a single
GPU; the harness, pinned environment (\code{transformers 5.11.0}), pre-registered decision rules, and
a record of what did and did not replicate are in the repository, alongside a longer version
with confidence intervals and the full audit.

\section{Conclusion}

Every repetition penalty is a workaround: an inference-time intervention correcting behavior the
model itself gets wrong. Workarounds differ in their sharp edges. The default multiplicative
operator branches on the logit zero-point, a coordinate training sets only incidentally; in
consequence, a fixed setting is a different operation on every model, equivalent gauges of one
model decode and cost differently at the same $\thp$, and routine strengths corrupt structured
output. Tuning per model and per task, the standard advice, presupposes that these edges are
known. We found one prior observation of the mechanism (Section~\ref{sec:stacks}) and no
documentation of either consequence. They belong to this
operator, not to repetition control in general: the subtractive penalties avoid them by
construction, though not every stack exposes one (HuggingFace's \code{generate} ships only the
multiplicative form), and the normalized variant of Section~\ref{sec:fix} removed A1 in every
measurement we ran, at matched suppression as well as matched $\thp$, and A2 at matched $\thp$, at
no detected quality cost. The calibration map states the trade: the normalized
operator is well-defined at every strength it can reach, and on all but one model measured, the
multiplicative operator's strongest settings, where the corruption occurs, lie beyond that
reach.

\bibliographystyle{plainnat}
\bibliography{refs}

\end{document}